\pdfoutput=1

\documentclass[11pt]{article}

\usepackage[preprint]{acl}

\usepackage{times}
\usepackage{latexsym}
\usepackage{colortbl}
\usepackage{xcolor}
\usepackage{array}
\usepackage{fontawesome5}
\usepackage{arydshln}
\usepackage[multiple]{footmisc}

\usepackage{booktabs}
\usepackage{subcaption}
\usepackage{authblk}
\usepackage{multirow}

\usepackage{amsmath}

\usepackage[T1]{fontenc}

\usepackage[utf8]{inputenc}

\usepackage{microtype}

\usepackage{inconsolata}

\usepackage{graphicx}
\usepackage{dingbat}

\usepackage{xspace}
\usepackage{pifont}
\usepackage{tablefootnote}
\usepackage{microtype}
\usepackage{tikz}
\usepackage{enumitem}

\DeclareMathOperator*{\argmax}{arg\,max}

\newcommand{\cmark}{{\color{black}\ding{51}}}
\newcommand{\xmark}{{\color{black}\ding{55}}}

\title{An Empirical Comparison of Generative Approaches \\for Product Attribute-Value Identification}

\author[$\ast$]{\textbf{Kassem Sabeh}}
\author[$\dagger$]{\textbf{Robert Litschko}}
\author[$\ddagger$]{\textbf{Mouna Kacimi}}
\author[$\dagger$]{\textbf{Barbara Plank}}
\author[$\ast$]{\textbf{Johann Gamper}}

\affil[$\ast$]{Free University of Bozen-Bolzano, Italy\\ \texttt{\{ksabeh, jgamper\}@unibz.it}}
\affil[$\dagger$]{LMU Munich, Germany\\ \texttt{\{robert.litschko, b.plank\}@lmu.de}}
\affil[$\ddagger$]{Wonder Technology Srl, Italy\\ \texttt{mouna@wonderflow.ai}}

\begin{document}

\maketitle
\begin{abstract}
Product attributes are crucial for e-commerce platforms, supporting applications like search, recommendation, and question answering. The task of Product Attribute and Value Identification (PAVI) involves identifying both attributes and their values from product information. In this paper, we formulate PAVI as a generation task and provide, to the best of our knowledge, the most comprehensive evaluation of PAVI so far. We compare three different attribute-value generation (AVG) strategies based on fine-tuning encoder-decoder models on three datasets. Experiments show that end-to-end AVG approach, which is computationally efficient, outperforms other strategies. However, there are differences depending on model sizes and the underlying language model. The code to reproduce all experiments is available at: \url{https://github.com/kassemsabeh/pavi-avg}
\end{abstract}

\section{Introduction}

\begin{figure}
\centering
\includegraphics[width=0.8\columnwidth]{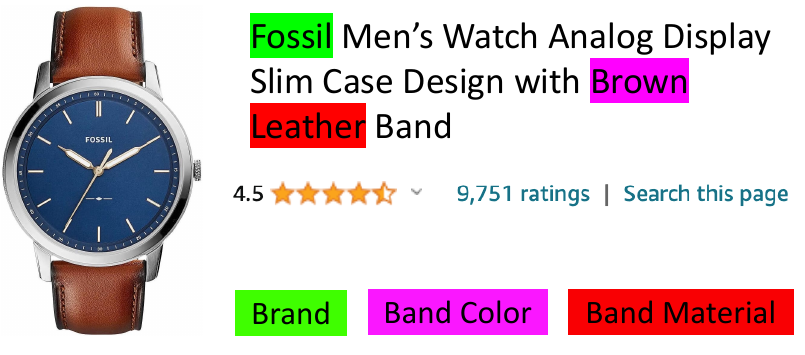}
\caption{An example of a product title with tagged attribute-value pairs.}
\label{fig:example}
\end{figure}

Product attributes are a crucial component of e-commerce platforms, facilitating applications such as product search \cite{chen2023generate}, product recommendation \cite{truong2022ampsum}, and product-related question answering \cite{deng2023product}. They provide useful details about product features, enabling customers to compare products and make informed purchasing decisions. Product attribute and value identification (PAVI) refers to the task of identifying both the attributes and their corresponding values from an input context, such as a product title or description. For example, given the product title "Fossil Men’s Watch Analog Display Slim Case Design with Brown Leather Band" (see Figure \ref{fig:example}), a model should identify the attributes \emph{Brand}, \emph{Band Color}, and \emph{Band Material}, with the corresponding values \emph{Fossil}, \emph{Brown}, and \emph{Leather}.

Most existing work focuses on product attribute-value extraction (PAVE) \cite{zheng2018opentag,xu2019scaling,wang2020learning,yang2022mave}, which extracts the value of a \textit{given} attribute from the input context. Despite extensive research on PAVE \cite{blume2023generative,yang2023mixpave,brinkmann2023product}, PAVI is a more realistic and complex task since it requires the attribute to be generated and not assumed to be part of the input. While recent studies have explored generative models for PAVI, these efforts are limited in scope and often lack comprehensive evaluation across different datasets and settings \cite{roy2024exploring,shinzato2023unified}. Moreover, existing work focus primarily on end-to-end models without exploring alternative generative strategies. Consequently, it remains unclear which types of PAVI models are effective in practice, as comprehensive experiments and comparisons are lacking.

In this paper, we address these gaps by proposing three generative approaches for PAVI and conducting a comprehensive evaluation across multiple datasets. Inspired by recent advancements on question and answer generation methods \cite{bartolo2021improving}, we compare between three strategies based on fine-tuning encoder-decoder language models such as T5 \cite{raffel2020exploring} and BART \cite{lewis2020bart}. Our proposed approaches are: (1) pipeline attribute-value generation (AVG), which decomposes the task into value extraction and attribute generation, and builds a separate model for each sub-task; (2) multitask AVG, which uses a single shared model that is trained on both sub-tasks; (3) end2end AVG, which uses a single model to generate the attribute-value pairs. We evaluate the performance of these approaches on three real-world product datasets: AE-110K, OA-mine, and MAVE. All the models and datasets are publicly released on HuggingFace\footnote{\url{https://huggingface.co/av-generation}} and available as a demo\footnote{\url{https://bit.ly/4bWFjNV}}.

\section{Related Work}

\begin{table}
\centering
\resizebox{\columnwidth}{!}{
\begin{tabular}{c|ccc|ccc|c}
\hline
 & Pipe. & Multi. & E2E& AE-110k & OA-MINE & MAVE & Open \\ \hline

\citet{shinzato2023unified} & \xmark & \xmark & \cmark & \xmark & \xmark & \cmark & \xmark \\
\citet{roy2024exploring} & \xmark & \xmark & \cmark & \cmark & \xmark & \xmark & \xmark \\\hline
Ours & \cmark & \cmark & \cmark & \cmark & 
\cmark &\cmark & \cmark \\ \hline

\end{tabular}
}
\caption{Comparison between our work and prior studies for generative-based PAVI.}
\label{tb:contributions}
\end{table}

Most existing approaches for attribute-value extraction use sequence tagging \cite{huang2015bidirectional,xu2019scaling,yan2021adatag,zheng2018opentag} or question answering \cite{wang2020learning,yang2022mave,ding2022ask,hu2022fusing,sabeh2022cave,yang2023mixpave} methods. However, such approaches carry closed-world assumption, as they require the set of attributes as inputs to extract the corresponding values. More recently, researchers have explored the capabilities of generative models to tackle the PAVI task, in an open-world setting. \citet{roy2024exploring} proposed a generative framework for joint attribute and value extraction. They conduct experiments on the AE-110k dataset and show that the generative approaches surpass question-answering based methods. \citet{shinzato2023unified} fine-tune a pre-trained T5 generative model \cite{raffel2020exploring} to decode a set of target attribute-value pairs from the input product text of the MAVE dataset \cite{yang2022mave}. They show that the generative approach outperforms extraction and classification-based methods \cite{chen2022extreme}. 

However, all above studies utilize an end-to-end generative approach. They did not explore other generative strategies for attribute-value identification (i.e., pipeline and multi-task). In addition, these approaches are not comparable as they are different in terms of datasets, settings, and evaluation metrics. Finally, none of the above proposed models have been made publicly available.
In this work, we propose three generative approaches for PAVI and empirically compare them on three real-world datasets. We summarize how our approach differs from prior work in Table \ref{tb:contributions}. As can be seen, 
we evaluate in total all approaches across three datasets.


\section{Proposed Methods}

\begin{figure}
\centering
\includegraphics[width=0.9\columnwidth]{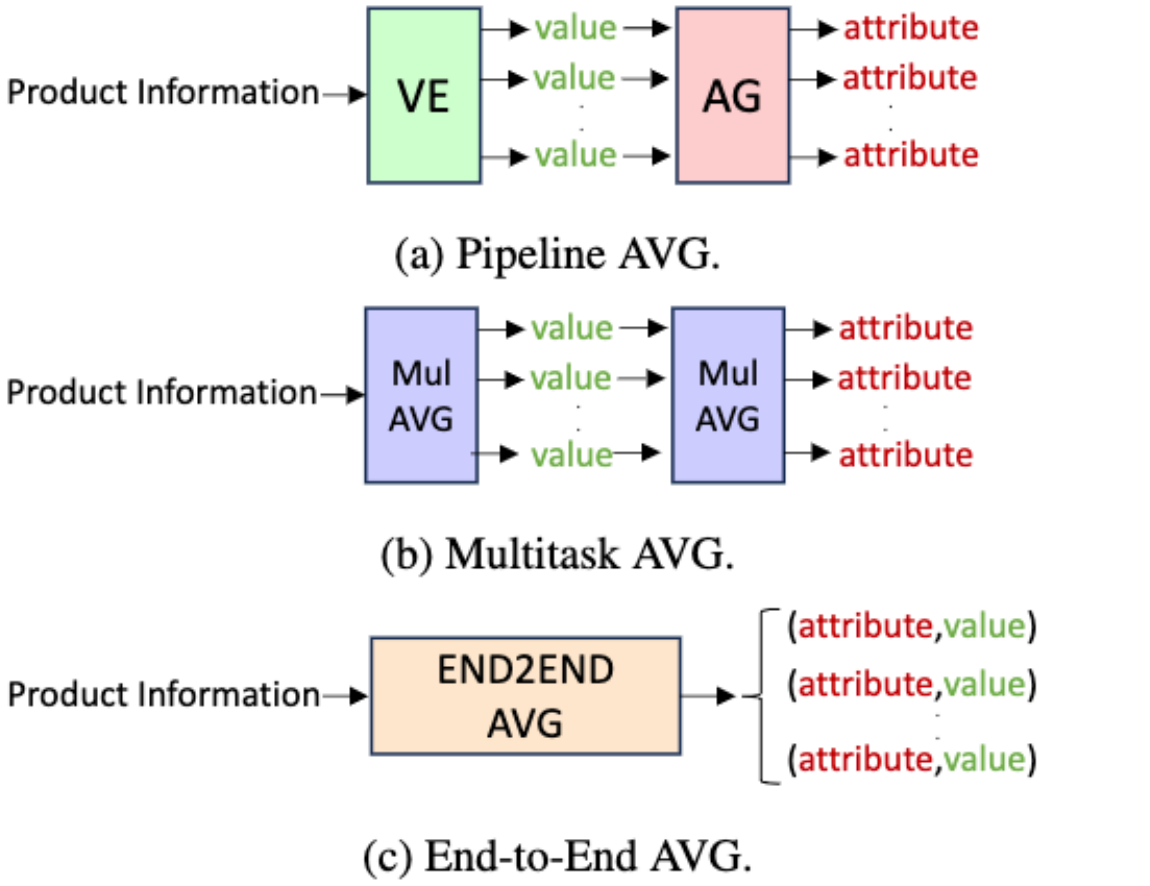}
\caption{Overview of the proposed AVG approaches.}
\label{fig:approaches}
\end{figure}

Given an input product data (title or description) $x=\{x_1, x_2, \ldots,x_{|x|}\}$, attribute-value generation aims to generate attribute-value pairs $\mathcal{Q}_{x}$ related to the information in $x$:
\begin{equation}
    \mathcal{Q}_{x} = \{(a^1,v^1),(a^2,v^2),(a^3,v^3), \ldots\}
\end{equation}
For instance, if $x$={"Fossil",\ldots,"Band"}, then $\mathcal{Q}_{x}$ = {("Brand","Fossil"), ("Band Color","Brown"), ("Band Material","Leather")}. 

We formulate the attribute-value identification problem as an attribute-value generation (AVG) task and propose three approaches based on fine-tuning language models, as depicted in Figure \ref{fig:approaches}.

\subsection{Pipeline AVG}


The AVG task can be decomposed into two simpler sub-tasks, value extraction (VE), and attribute generation (AG). The VE model $P_{ve}$ first generates the value candidate $\tilde{v}$ from $x$. Then, the AG model $P_{ag}$ generates an attribute $\tilde{a}$ whose value is $\tilde{v}$ in the input $x$. The VE and AG models can be trained independently on a product dataset consisting of the triplet $(x, a, v)$ by maximizing the conditional log likelihood of:
\begin{align} 
\tilde{v}=\argmax_v P_{ve}(v\mid x) \\
\tilde{a}=\argmax_a P_{ag}(a\mid x,v)
\end{align}
%
In practice, the VE model input is $[x_1, x_2, \ldots x_{|x|}]$,
where $x_i$ is the i-th token of the product input $x$ and ${|\cdot|}$ represents the number of tokens in the sequence. 
The input to the AG model takes the value into account by highlighting it inside the input. Specifically, following previous work \cite{chan2019recurrent,ushio2023empirical}, we introduce a highlight token \texttt{<hl>} to take the value into account:
\begin{equation*}
[x_1, \ldots,\texttt{<hl>}, v_1, \ldots,v_{|v|},\texttt{<hl>}, \ldots x_{|x|}]
\end{equation*}
where $v_i$ is the i-th token of $v$. At inference, we simply replace the gold value $v$ of the AG model by the prediction from the VE model, and run the inference over the product context $x$. For example, if the VE model extracts "Leather" from the input
$x$, we highlight "Leather" and feed it to the AG model as:  ["Fossil",\ldots,\texttt{<hl>},"Leather",\texttt{<hl>},\ldots,"Band"].  Thus, the pipeline approach generates at most one attribute-value pair per product context 
$x$.

To allow the pipeline approach to generate multiple attribute-value pairs, we can convert the values into a flattened sentence $y$, and fine-tune a sequence-to-sequence model to generate $y$ from $x$. Formally, we define a function $\mathcal{L}$ that maps $\mathcal{Q}_{x}$ to a sentence as:
\begin{align}
\mathcal{L}(\mathcal{Q}_{x})="v_1 | v_2 | v_3 \ldots".
\end{align}
%
In this case, the VE model generates a set of possible values, and for each value we run the AG model to obtain a set of attribute-value pairs.

\subsection{Multitask AVG}

Instead of training two separate generative models for each sub-task, we can instead use a single shared model that is fine-tuned in a multi-task learning setting. Namely, we mix the training instances for the VE and AG tasks together, and randomly sample a batch at each iteration of seq2seq fine-tuning. We distinguish each task by adding a prefix to the beginning of the input text. Namely, we add $\texttt{extract value}$ for the VE task, and $\texttt{generate attribute}$ for the AG task. 


\subsection{End2End AVG}



Instead of breaking the AVG task into two sub-tasks, we can directly model it by transforming the target attribute-value pairs to a flattened sentence $z$, and fine-tune a seq2seq model to directly generate the $z$ from $x$. We define a function $\mathcal{T}$ that maps the target $\mathcal{Q}_{x}$ to a sentence as:
\begin{align}
\mathcal{T}(\mathcal{Q}_{x})="\{t(a^1,v^1) | t(a^2,v^2) | \ldots\}". \\
t(a,v) = "\texttt{attribute}:\{a\},\texttt{value}:\{v\}"
\end{align}
We use the template $t$ to textualize the attribute-value pairs and separate them using a separator \texttt{|}. 
The end2end AVG model $P{avg}$ is optimized by maximizing the conditional log-likelihood:
\begin{align}
\tilde{z}=\argmax_z P_{avg}(z\mid x)
\end{align}

\section{Experimental Settings} \label{sec: settings}

\noindent \textbf{Datasets.}
We use three real-world datasets.

\begin{itemize}[leftmargin=*,noitemsep,nolistsep]
    \item AE-110K \cite{xu2019scaling}: This dataset contains tuples of product titles, attributes, and values from AliExpress Sports \& Entertainment category. Instances with NULL values are removed, resulting in 39,505 products with 2,045 unique attributes and 10,977 unique values.
    \item MAVE \cite{yang2022mave}: This is a large and diverse dataset complied from the Amazon Review Dataset \cite{ni2019justifying}. We remove negative examples from the MAVE dataset, where there are no values for the attributes. The final dataset contains around 2.9M attribute-value annotations from 2.2M cleaned Amazon products.
    \item OA-Mine \cite{zhang2022oa}: We use the human-annotated dataset, which contains 1,943 product data from 10 product categories. No further processing is applied to this dataset.
\end{itemize}

\noindent 
We randomly split all datasets in train:val:test = 8:1:1. The splits are stratified by product category. Appendix \ref{app:datasets} shows statistics of the three datasets.

\begin{table*}[tb]
\centering
\footnotesize
\begin{tabular}{cc|ccc|ccc|ccc}
\hline
\multicolumn{2}{c}{\multirow{2}{*}{Approach}} & \multicolumn{3}{c}{AE-110k} & \multicolumn{3}{c}{OA-Mine} & \multicolumn{3}{c}{MAVE} \\ \cline{3-11} 
\multicolumn{2}{c}{} & $P$ & $R$ & $F_1$ & $P$ & $R$& $F_1$ & $P$ & $R$ & $F_1$ \\ \hline
\multicolumn{1}{c}{\multirow{4}{*}{T5 Small}} & Pipeline & 94.61 & 70.62 & 80.88 & 69.85 & 76.10 & 72.84 & 91.51 & 89.60 & 90.55 \\
\multicolumn{1}{c}{} & Multitask & \underline{94.94} & \underline{73.00} & \underline{82.53} & \underline{73.70} & \underline{79.46} & \underline{76.48} & \underline{94.88} & \underline{92.87} & \underline{93.86} \\
\multicolumn{1}{c}{} & End2End & 94.07 & 70.45 & 80.56 & 65.12 & 49.57 & 56.29 & 90.22 & 90.29 & 90.25 \\ \cdashline{2-11}
\multicolumn{1}{c}{} & Ensemble & 93.25 & 79.74 & 85.97 & 72.38 & 86.24 & 78.71 & 91.49 & 95.82 & 93.60 \\ \hline
\multicolumn{1}{c}{\multirow{4}{*}{T5 Base}} & Pipeline & 94.93 & 73.74 & 83.01 & 78.82 & 87.46 & 82.92 & 92.10 & 91.52 & 91.80 \\
\multicolumn{1}{c}{} & Multitask & 95.50 & \underline{74.55} & \underline{83.74} & \underline{79.83} & \underline{89.22} & \underline{84.26} & \underline{96.19} & \underline{\textbf{94.10}} & \underline{95.14} \\
\multicolumn{1}{c}{} & End2End & \underline{\textbf{95.61}} & 74.44 & 83.71 & 79.63 & 82.36 & 80.98 & 90.31 & 91.01 & 90.65 \\ \cdashline{2-11}
\multicolumn{1}{c}{} & Ensemble & 93.82 & 91.27 & 87.10 & 79.11 & 94.58 & 86.15 & 91.72 & 96.76 & 94.18 \\ \hline
\multicolumn{1}{c}{\multirow{4}{*}{T5 Large}} & Pipeline & 94.15 & 73.83 & 82.76 & 78.76 & 88.70 & 83.43 & 92.34 & 91.32 & 91.82 \\
\multicolumn{1}{c}{} & Multitask & 94.89 &  69.73 & 80.38 & 81.43 & \underline{\textbf{90.30}} & 85.63 & 96.21 & 92.51 & 94.32 \\
\multicolumn{1}{c}{} & End2End & \underline{95.21} & \underline{\textbf{75.62}} & \underline{\textbf{84.29}} & \underline{\textbf{82.69}} & 90.20 & \underline{\textbf{86.28}} & \underline{\textbf{96.39}} & \underline{94.01} & \underline{\textbf{95.19}} \\ \cdashline{2-11}
\multicolumn{1}{c}{} & Ensemble & 92.75 & 81.57 & 86.80 & 80.63 & 95.79 & 87.56 & 91.95 & 96.89 & 94.36 \\ \hline
\multicolumn{1}{c}{\multirow{4}{*}{BART Base}} & Pipeline & 95.00 & 70.73 & 81.09 & 76.25 & 85.05 & 80.41 & \underline{91.20}  & 89.87 & \underline{90.53} \\
\multicolumn{1}{c}{} & Multitask & \underline{95.07} & \underline{71.66} & \underline{81.72} & \underline{78.78} & \underline{87.27} & \underline{82.81} & 89.92 & \underline{90.74} & 90.33  \\
\multicolumn{1}{c}{} & End2End & 83.33 & 51.86 & 63.93 & 50.85 & 39.04 & 44.17 & 79.46 & 87.40 &  83.24 \\ \cdashline{2-11}
\multicolumn{1}{c}{} & Ensemble & 92.71 & 78.82 & 85.21 & 77.30 & 92.16 & 84.08 & 90.53 & 96.20 & 93.27 \\ \hline
\multicolumn{1}{c}{\multirow{4}{*}{BART Large}} & Pipeline & \underline{94.81} & 68.40 & 79.47 & 78.18 & 86.84 & 82.29 & \underline{92.13} & 90.21 &  \underline{91.16} \\
\multicolumn{1}{c}{} & Multitask & 94.42 & \underline{72.52} & \underline{82.04} & \underline{78.62} & \underline{87.96} & \underline{83.03} & 90.47 & \underline{91.41} & 90.94 \\
\multicolumn{1}{c}{} & End2End & 63.02 & 46.66 & 53.62 & 48.83 & 37.24 & 42.26 & 77.29 & 86.45 & 81.61 \\ \cdashline{2-11}
\multicolumn{1}{c}{} & Ensemble & 92.47 & 79.10 & 85.26 & 77.86 & 93.90 & 85.14 & 91.34 & 96.47 & 93.85 \\ \hline
\end{tabular}
\caption{Evaluation results of different attribute-value generation methods. The best score among the approaches for each language model is underlined, and the best result in each dataset across all models is in boldface.}
\label{tb:results}
\end{table*}

\noindent \textbf{Base Models.} For all approaches (pipeline, multitask, and end2end), we experiment with the base language models T5 \cite{raffel2020exploring} and BART \cite{lewis2020bart}. We also compare between the model weights \texttt{t5-\{small,base,large\}} and \texttt{facebook/bart-\{base,large\}} from HuggingFace\footnote{\url{https://huggingface.co/}}\footnote{See Appendix \ref{app:hyper-parameters} for Hyper-parameter details.}.

\noindent \textbf{Evaluation Metrics.} Following previous works \cite{yang2022mave,shinzato2023unified}, we use precision $P$, recall $R$, and $F_1$ score as evaluation metrics. The datasets may contain missing attribute-value pairs that the model might generate. To reduce the impact of such missing attribute-value pairs \cite{shinzato2023unified}, we discard predicted attribute-value pairs if there are no ground truth labels for the generated attributes.



 
\section{Results}


Table \ref{tb:results} provides the main results. In addition to the three approaches (i.e., pipeline, multitask, and end2end), we also provide an ensemble model that combines the generated attribute-value pairs from these approaches. Overall, T5 large (end2end) achieves the best scores across the three datasets. Additionally, the multitask approach exhibits commendable performance, often ranking the second best. There are several interesting observations in Table \ref{tb:results}. First, while the end2end approach generally excels, there are instances where the pipeline or multitask approach outperforms it, especially with smaller model sizes. For example, for T5 small on the OA-Mine dataset, the multitask approach outperforms end2end with an $F_1$ score of 76.48 compared to 56.29. 
By analyzing the errors, we found that the end2end approach makes more errors in detecting attributes, which the multitask approach mitigates. This improvement is mainly because the multitask approach has been specifically trained on the task of attribute generation. Second, the influence of model size on performance is evident, with larger models generally achieving better results across all approaches. For instance, T5 base and T5 large consistently outperform T5 small across all datasets and approaches. This trend is also seen with BART models. Third, among the AVG approaches, T5 consistently works better with the end2end AVG, while BART is not well-suited when used end2end. A possible explanation is that T5 has observed sentences with structured observation due to its multitask pre-training objective, while BART did not encounter such training instances as it was trained only on a denoising sequence-to-sequence objective. Finally, there are notable differences in performance across the datasets. For instance, the MAVE dataset sees higher overall $F_1$ scores compared to AE-110k and OA-Mine datasets. The higher results on the MAVE dataset can be attributed to its uniform annotation process using an ensemble of models, unlike the more varied human annotations in AE-110k and OA-Mine\footnote{See Appendix \ref{app: cross-dataset} for cross-dataset evaluation.}.

Ensemble models, which combine the generated attribute-value pairs across the three approaches, consistently improve results. For instance, in AE-110k, ensembling trades off a small amount of precision for substantial gains in recall, while in OA-Mine, precision remains stable with improved recall. In general, ensembling helps to identify more attributes and therefore enhances the $F_1$ score by increasing the recall. However, it slightly reduces precision due to challenges in extracting accurate values for these new attributes.

\section{Conclusion}
In this paper, we formalized PAVI as an attribute-value generation task and established three different AVG approaches. Using T5 and BART base models, we conducted experiments on three benchmark product datasets. Our evaluation demonstrates that end2end AVG, which generates attributes and values simultaneously, is generally more reliable. However, pipeline or multitask approach can offer advantages, particularly for smaller models and when using language models like BART.

\section*{Limitations}

Our study has two main limitations. First, the datasets used in our experiments do not have standard splits. We randomly split the datasets as discussed in Section \ref{sec: settings}, but we have provided the exact data splits in our repository to ensure reproducibility and comparability. Second, the evaluation measures employed do not penalize over-generated attribute-value pairs. We assume that the datasets do not have all possible annotations, so the generative models might correctly identify new attribute-value pairs. However, in our evaluation, we discard these newly generated attribute-value pairs. As future work, we plan to develop methods for the automatic evaluation of newly generated attribute-value pairs.


\bibliography{custom}

\appendix

\begin{table}[tb]

\centering
\resizebox{\columnwidth}{!}{
\begin{tabular}{cccc}

\hline
\textbf{Counts} & AE-110K & OA-Mine & MAVE \\ \hline

\# products & 39,505 & 1,943 & 2,226,509\\
\# attribute-value pairs & 88,915 & 11,008 & 2,987,151 \\
\# unique categories & 10 & 10 & 1,257 \\
\# unique attributes & 2,045 & 51 & 705 \\
\# unique values & 10,977 & 5,201 & 79,199 \\ \hline

\end{tabular}
}
\caption{Statistics of AE-110K, OA-Mine, and MAVE datasets.}
\label{tb:statistics}
\end{table}

\begin{table}
\centering
\resizebox{\columnwidth}{!}{
\begin{tabular}{llccc}
\toprule
\multicolumn{2}{c}{Approach} & Epochs & LR & Batch Size \\ 
\midrule
\multirow{4}{*}{T5 small} & Pipeline (VE) & 9 & $5e^{-5}$ & 128 \\
                          & Pipeline (AG) & 11 & $5e^{-5}$ & 128 \\
                          & Multitask     & 16 & $5e^{-4}$ & 256 \\
                          & End2End       & 18 & $5e^{-4}$ & 256 \\ 
\midrule
\multirow{4}{*}{T5 base}  & Pipeline (VE) & 8 & $5e^{-4}$ & 64 \\
                          & Pipeline (AG) & 7 & $5e^{-4}$ & 64 \\
                          & Multitask     & 8 & $5e^{-4}$ & 128 \\
                          & End2End       & 11 & $5e^{-4}$ & 64 \\
\midrule
\multirow{4}{*}{T5 large}  & Pipeline (VE) & 6 & $5e^{-5}$ & 128 \\
                          & Pipeline (AG) & 5 & $5e^{-4}$ & 64 \\
                          & Multitask     & 5 & $1e^{-4}$ & 64 \\
                          & End2End       & 8 & $1e^{-4}$ & 64 \\
\midrule
\multirow{4}{*}{BART base}  & Pipeline (VE) & 5 & $5e^{-5}$ & 64 \\
                          & Pipeline (AG) & 4 & $1e^{-4}$ & 128 \\
                          & Multitask     & 4 & $1e^{-4}$ & 64 \\
                          & End2End       & 6 & $5e^{-4}$ & 128 \\
\midrule
\multirow{4}{*}{BART large}  & Pipeline (VE) & 6 & $5e^{-5}$ & 64 \\
                          & Pipeline (AG) & 4 & $5e^{-5}$ & 128 \\
                          & Multitask     & 3 & $1e^{-5}$ & 64 \\
                          & End2End       & 7 & $1e^{-5}$ & 64 \\
\bottomrule
\end{tabular}
}
\caption{Hyper-parameter details, including number of training epochs, learning rate (LR), and batch size, for different AVG approaches.}
\label{tb:hyper-parameters}
\end{table}

\section{Datasets}
\label{app:datasets}

Table \ref{tb:statistics} shows the statistics of the three datasets: AE-110K \cite{xu2019scaling}, OA-Mine \cite{zhang2022oa}, and MAVE \cite{yang2022mave}.

\section{Hyper-Parameters}

\label{app:hyper-parameters}
All models are implemented using Pytorch and are trained on NVIDIA Tesla A100 GPUs. We use the validation set of the datasets to select the optimal hyper-parameters for all models, while we report our final results on the test set. During training, optimization is performed using Adam \cite{diederik2014adam} optimizer. We perform early stopping if there is no improvement in the loss on the validation set for 3 epochs. The maximum input length is fixed at 512. The maximum output length is 256 for the end2end models and 64 for the others. The details of all hyper-parameters for each approach are reported in Table \ref{tb:hyper-parameters}.

\section{Model Comparison}

\begin{table}
    \centering
    \resizebox{\columnwidth}{!}{%
    \begin{tabular}{lcccc}
    \hline
        Approach & Training Cost & Inference Cost & Memory & Generated AV \\ \hline
        Pipeline & 3.9$\times$ & 2.7$\times$ & 2$\times$ & 2.3$\times$ \\
        Multitask & 2.5$\times$ & 2.7$\times$ & 1$\times$ & 2.3$\times$ \\
        End2End & 1$\times$ & 1$\times$ & 1$\times$ & 1$\times$ \\ \hline
    \end{tabular}
    }
        \caption{Training cost, inference cost, memory requirements, and number of generated attribute-value pairs (Generated AV) of the three proposed AVG approaches, normalized to the End2End approach. The comparison is performed for \texttt{T5-large}. Generated AV are averaged across the three datasets.}
    \label{tb:costs}
\end{table}

In addition to performance, computational cost and usability are crucial factors when selecting an AVG approach. Table \ref{tb:costs} details the training cost, inference cost, memory requirements, and number of generated attribute-value pairs for the pipeline, multitask, and end2end approaches. The end2end approach is the most efficient during training due to its single integrated model. In contrast, the pipeline approach has the highest training cost because it requires training two separate models for its sequential processing stages. The multitask approach falls in between, as it uses a shared model, reducing redundancy and thus lowering the training cost compared to the pipeline approach. For inference, end2end AVG is the fastest as it can generate the attribute-value pairs in a single pass. Both pipeline and multitask approaches are slower since they handle each task independently, and a single prediction requires two steps: value extraction and attribute generation. Regarding memory requirements, both end2end and multitask AVG employ a single model, while pipeline AVG uses two separate models, effectively doubling the memory footprint. 
While the end2end approach is the most efficient overall, minimizing both training and inference costs, the pipeline and multitask approaches can generate a larger number of attribute-value pairs on average. Additionally, the pipeline and multitask approaches offer flexibility for separate processes, as they can perform value extraction or attribute generation sub-tasks independently.

\begin{figure*}[t]
\centering
\begin{subfigure}[b]{1\columnwidth}
  \includegraphics[width=\columnwidth]{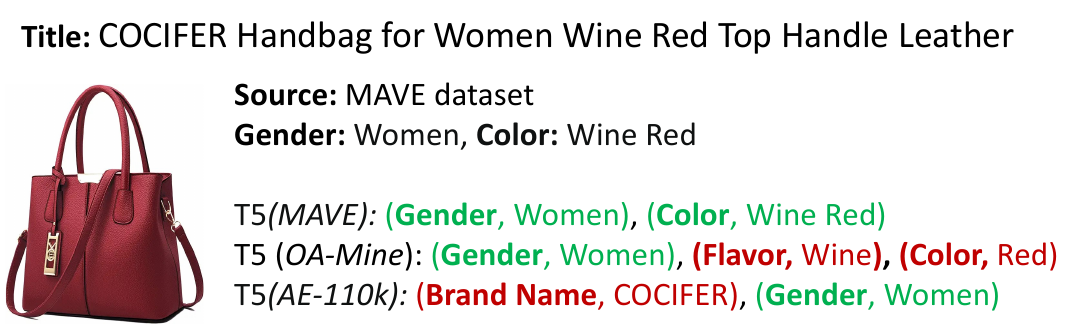}
  \caption{Handbag product from MAVE dataset.}
  \label{fig:example_a} 
\end{subfigure}
   ~ 
\begin{subfigure}[b]{1\columnwidth}
  \includegraphics[width=\columnwidth]{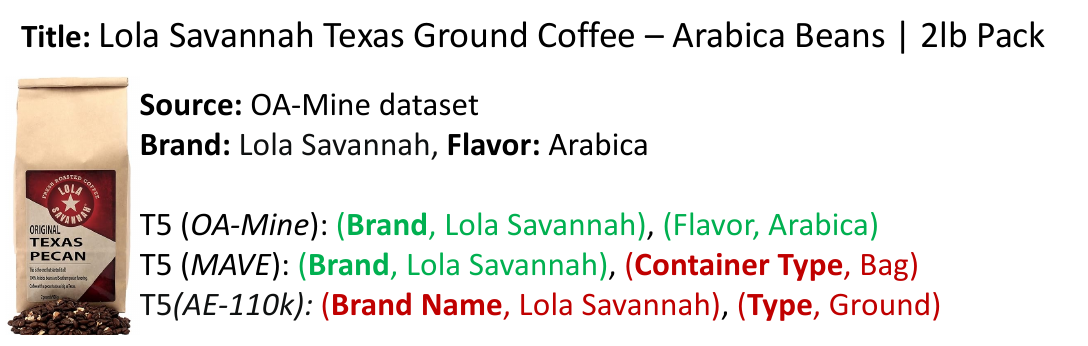}
  \caption{Coffee product from OA-Mine dataset.}
  \label{fig:example_b}
\end{subfigure}

\caption{Examples of cross-domain attribute-value identification. Correct predictions are highlighted in green, and wrong ones are highlighted in red. In the first example, the T5 model trained on OA-Mine incorrectly predicts food-related attributes, showing domain bias. While the in-domain T5 model, trained on MAVE dataset, correctly identifies all attribute-value pairs. In the second example, both T5 models trained on MAVE and AE-110K (cross-domain), fail to identify the \emph{Flavor} attribute.}
\label{fig:examples}
\end{figure*}

\section{Cross-dataset Evaluation}
\label{app: cross-dataset}
\begin{table}
    \centering
\begin{tabular}{cccc}
\hline
         & AE-110K & OA-Mine & MAVE \\
         \hline
         AE-110k & \textbf{84.29} & 1.58 & 6.40 \\
         OA-Mine & 16.42 & \textbf{86.28} & 4.31 \\
         MAVE & 6.42 & 3.18 & \textbf{95.19} \\
         \hline
    \end{tabular}
    \caption{$F_1$ scores of cross-dataset predictions of \texttt{T5-large} end2end model.}
    \label{tb:cross-dataset}
\end{table}

Table \ref{tb:cross-dataset} presents the $F_1$ scores of the T5-large end2end model evaluated on cross-dataset predictions. We chose the T5-large end2end model because it demonstrated the best in-domain performance, as shown in Table \ref{tb:results}. The models exhibit high performance when evaluated on the same dataset used for training, with $F_1$ scores of 84.29, 86.28, and 95.19 on AE-110K, OA-Mine, and MAVE respectively. This indicates the models' strong ability to fit the training data. However, there is a notable drop in performance when the models are tested on different datasets. For instance, when the model trained on AE-110K is evaluated on OA-Mine and MAVE, the $F_1$ scores drop to 1.58 and 6.40 respectively. Similarly, models trained on OA-Mine and MAVE also show reduced performance on other datasets, with $F_1$ scores as low as 4.31 and 3.18. These results highlight a significant challenge in the model's ability to generalize across different datasets. The poor cross-dataset performance can be attributed to the different attribute names and categories/domains present in each dataset, which the model struggles to generate. Figure \ref{fig:examples} provides two examples illustrating these challenges. In the first example of a Handbag product from the MAVE dataset, the model trained on OA-Mine predicts (Gender, "Woman"), (Flavor, "Wine"), and (Color, "Red"), while the model trained on AE-110k predicts (Brand Name, "COCIFIER") and (Gender, "Women"). This example demonstrates the domain bias of the model trained on OA-Mine, which includes food-related items. The model erroneously identifies the attribute "Flavor" for the value "Wine", a food-related attribute, when applied to a fashion product.

In the second example of a coffee product from the OA-Mine dataset, the model trained on MAVE predicts (Brand, "Lola Savannah") and (Container Type, "Bag"), while the model trained on AE-110k predicts (Brand Name, "Lola Savannah") and (Type, "Ground"). The differences in attribute names across datasets, such as "Brand" versus "Brand Name", lead to incorrect predictions. Additionally, since the MAVE and AE-110k datasets do not include products from food categories, they fail to identify the "Flavor" attribute, which is specific to the OA-Mine dataset.

\end{document}